%% file: main.tex
\documentclass[11pt,a4paper]{article}

\usepackage[utf8]{inputenc}
\usepackage[T1]{fontenc}
\usepackage{geometry}
\usepackage{parskip}      
\usepackage{xcolor}
\usepackage{amsmath}
\usepackage{abstract}
\usepackage{graphicx}     
\usepackage[hyphens]{url}
\usepackage[hidelinks]{hyperref}

\usepackage[natbibapa]{apacite}
\usepackage{listings}
\definecolor{codegreen}{rgb}{0,0.6,0}
\definecolor{codegray}{rgb}{0.5,0.5,0.5}
\definecolor{codepurple}{rgb}{0.58,0,0.82}
\definecolor{backcolour}{rgb}{0.95,0.95,0.92}
\lstdefinestyle{mystyle}{
    backgroundcolor=\color{backcolour},
    commentstyle=\color{codegreen},
    keywordstyle=\color{magenta},
    numberstyle=\tiny\color{codegray},
    stringstyle=\color{codepurple},
    basicstyle=\ttfamily\footnotesize,
    breakatwhitespace=false,
    breaklines=true,
    captionpos=b,
    keepspaces=true,
    numbers=left,
    numbersep=5pt,
    showspaces=false,
    showstringspaces=false,
    showtabs=false,
    tabsize=2
}
\usepackage{booktabs}
\usepackage{array}
\usepackage{tabularx}
\usepackage{multirow}
\usepackage{longtable}
\usepackage{threeparttable}
\usepackage{threeparttablex}
\usepackage{ragged2e}
\usepackage{float}        

\newcolumntype{P}[1]{>{\RaggedRight\arraybackslash}p{#1}}
\newcolumntype{Y}{>{\RaggedRight\arraybackslash}X}
\title{\textbf{Retrieval-Augmented Generation of Pediatric Speech-Language Pathology vignettes: \\A Proof-of-Concept  Study }}
\author{
    Yilan Liu\thanks{University of Redlands, Department of Communication Sciences and Disorders, Truesdail Clinic Center. Corresponding author: \texttt{yilan\_liu@redlands.edu}}
  }

\begin{document}

\maketitle
\input{1-abstract}

\input{2-introduction}

\input{3-methods}
\input{4-results}
\input{5-discussion}
\input{6-conclusion}

\setlength{\bibsep}{12pt} 
\bibliography{7-references}

\input{8-appendices}

\end{document}

%% file: 1-abstract.tex
\begin{abstract}
\textbf{Background: } Clinical vignettes are essential educational tools in speech-language pathology (SLP), yet manual creation is time-intensive, limiting educational and research applications. While general-purpose large language models (LLMs) demonstrate text generation capabilities, they lack domain-specific knowledge, exhibiting increased hallucinations and producing outputs requiring extensive expert revision. This study presents a proof-of-concept system integrating retrieval-augmented generation (RAG) with curated knowledge bases to generate pediatric SLP case materials.

\textbf{Method: }A multi-model RAG-based system was prototyped integrating curated domain knowledge with engineered prompt templates. The system supports five LLMs: three commercial models (GPT-4o, Claude 3.5 Sonnet, Gemini 2.5 Pro) and two open-source models (Llama 3.2, Qwen 2.5-7B). Seven test scenarios were systematically designed spanning diverse disorder types and grade levels. Generated cases underwent automated quality assessment using a multi-dimensional rubric evaluating structural completeness, internal consistency, clinical appropriateness, and IEP goal/session note quality on a 5-point scale.

\textbf{Results: } This proof-of-concept demonstrates technical feasibility of RAG-augmented generation for pediatric clinical vignettes in SLP across multiple LLM implementations. Commercial models showed marginal quality advantages, but open-source alternatives achieved acceptable performance for preliminary educational applications, suggesting potential for privacy-preserving institutional deployment. Integration of curated knowledge bases enables generation of content aligned with professional guidelines and documentation standards.

\textbf{Conclusions: }This proof-of-concept demonstrates technical feasibility of RAG-augmented generation for school SLP vignettes. Integration of curated knowledge bases enables generation of content aligned with professional guidelines and documentation standards. Extensive validation through expert review, student pilot testing, and psychometric evaluation is required before educational or research implementation. Applications may extend to clinical decision support system development, automated IEP goal generation tools, and clinical reflection training.


\end{abstract}
\pagebreak

%% file: 2-introduction.tex
\section{Introduction}

\subsection{Background}

Synthetic data, artificially generated data preserving statistical properties of real-world data without containing actual patient information, has demonstrated utility across healthcare domains. Synthetic data encompasses diverse forms including clinical vignettes (case scenarios), electronic health record simulations, medical imaging datasets, and pharmaceutical trial data \citep{giuffre2023harnessing}. Validation studies of synthetic electronic health records report fidelity exceeding 90\% for key clinical variables \citep{Chen2024Fairness}. In medical education, AI-generated clinical vignettes achieve 97\% accuracy after expert review \citep{Nakamura2024} with psychometric properties comparable to human-authored materials \citep{Sridharan2024}.

Speech-language pathologists (SLPs) have employed clinical vignettes for decades to develop clinical reasoning and evaluate decision-making \citep{Peabody2000}. However, unlike medicine and nursing where AI-generated materials have been extensively validated, SLP education has not adopted scalable synthetic data generation for clinical vignettes. Simulation-based learning in SLP commonly employs standardized patients (actors trained to portray clinical scenarios), video-based simulations, and part-task trainers \citep{Dudding2018}. Commercial simulation programs exist but may not be readily transferable between institutions due to mismatches in learning objectives, curricula, available resources, and infrastructure \citep{AlGhareeb2016}. 

This gap is particularly acute for school-based contexts requiring integration of American Speech-Language-Hearing Association (ASHA) clinical guidelines, IEP development frameworks under IDEA, state educational standards, school-based documentation requirements distinct from medical settings. Current vignette creation in SLP education remains predominantly ad hoc, relying on individual instructor effort without systematic frameworks, quality assurance, or scalable infrastructure \citep{Cowie2021}. This creates critical limitations constraining educational practice, research, and technology development.

Recent advances in artificial intelligence (AI), particularly large language models (LLMs), have demonstrated promising capabilities for generating clinical cases in medical education. Studies show that AI-generated cases can significantly accelerate case creation, improve diversity and cultural responsiveness, and provide virtually unlimited practice scenarios for students \citep{Bakkum2024}. A recent study reported that researchers generated sets of 30 diverse medical case vignettes in approximately 60 minutes using optimized prompts, a dramatic reduction from the hours or days typically required for manual case creation \citep{Bakkum2024}. However, these advances also reveal two fundamental challenges that must be addressed for effective application in specialized clinical domains like school-based speech-language pathology: framed prompts and domain-specific knowledge context.

The prompt challenge stems from the sophisticated input design required to guide general-purpose LLMs toward clinically appropriate outputs. Prompt engineering, the practice of designing structured input instructions that guide model outputs toward desired specifications \citep{White2023, Sahoo2024}, has demonstrated substantial improvements in output quality when systematically developed, including significant gains in consistency and reliability \citep{Wang2024}. Optimized prompts can reduce errors and increase clinical appropriateness in AI-generated medical vignettes \citep{Bakkum2024, Kim2025}. However, developing effective prompts requires technical understanding of AI model behavior, limitations, and response patterns. This technical expertise creates a significant barrier to scalable adoption. Individual clinicians experimenting with conversational AI must invest substantial time learning prompt engineering principles, iterating through trial-and-error refinement, and developing domain-specific prompt templates \citep{Kim2025}. Moreover, ad hoc manual prompting across different users produces inconsistent outputs, as variations in prompt phrasing, specificity, and structure yield substantial differences in generation quality \citep{Liu2023}. Pre-service SLPs and clinicians typically lack training in AI interaction strategies, making spontaneous high-quality prompt creation unrealistic without dedicated instruction.

The domain knowledge challenge arises from the gap between general-purpose LLMs' training and the specialized knowledge required for school-based SLP contexts. Conversational AI with general-purpose LLMs (e.g., ChatGPT, Claude.ai, Gemini) lack domain-specific knowledge essential for school-based SLP and may increase hallucinations, inaccurate or fabricated content stemming from inadequate domain-specific training data, limited exposure to specialized content, and insufficient knowledge coverage for rare conditions \citep{Nazi2025}. While biomedical LLMs trained on specialized corpora show improved understanding compared to general models \citep{Nazi2025}, domain-specific adaptations remain insufficient without external knowledge retrieval. School-based speech therapy simulations require synthesis of interconnected knowledge, such as ASHA practice guidelines, specific clinical knowledge, IEP frameworks and IDEA legal requirements, state educational standards integrated with speech-language goals, documentation requirements distinct from medical settings, and progress monitoring frameworks aligned with educational contexts. Without systematic integration of these knowledge bases, generated content cannot simulate real-world scenarios or meet professional standards. Evaluation of AI-generated SLP intervention plans found outputs rated "Needs Improvement" to "Meets Expectations" with considerable variability \citep{Kim2025}, and validation studies emphasize the necessity of expert review \citep{Nakamura2024}, indicating that even optimized models require human oversight.

Generating annual IEP goals exemplifies how both prompt engineering and clinical knowledge integration are essential for high-quality outputs. From a prompt engineering perspective, generating an appropriate IEP goal requires prompts specifying SMART criteria (Specific, Measurable, Achievable, Relevant, Time-bound), educational relevance, state standard alignment, measurement procedures, baseline data integration, and developmentally appropriate activities \citep{Hedin2018, Yell2017}. Constructing such comprehensive prompts demands knowledge of how to structure information for AI processing, which instructions take priority, and how to format complex multi-component requirements. Without standardized, validated prompt templates embedded in system architecture, each user must independently develop prompting strategies, leading to quality variability and limiting scalability. 

From a clinical perspective, school-based SLPs must produce SMART goals aligned with state standards and educationally relevant to curriculum access, requirements that differ substantially from medical settings. The novice SLPs and clinical fellows report difficulty with goal writing, often relying on generic goal banks containing poorly written, vague, non-measurable goals failing to individualize to student needs \citep{Rakap03042015}. Expert clinicians develop goal-writing skills through exposure to high-quality examples, supervisor feedback, and iterative refinement, internalizing patterns of effective goal structure, meaningful data collection, and documentation satisfying both clinical and compliance requirements. Conversational AI with general-purpose LLMs lacks curated collections of school-based SLP documentation exemplary, preventing realistic, high-quality generation without external knowledge sources.

\subsection{RAG: Grounding AI in Domain Knowledge}

Retrieval-augmented generation (RAG), combined with engineered prompt templates, addresses these fundamental limitations by: (1) grounding generation in authoritative domain knowledge through real-time retrieval from curated sources, and (2) encoding expert clinical and documentation knowledge into reusable prompt structures. Unlike fine-tuning requiring extensive retraining, RAG retrieves information at runtime, enabling access to current guidelines without model retraining \citep{Stroum2025}. RAG systems demonstrate superior performance over traditional approaches in clinical decision support, diagnostic assistance, and medical information extraction \citep{Chen2025}.

RAG addresses three key deficiencies. First, RAG reduces hallucinations by anchoring generation in verified documents rather than parametric knowledge alone \citep{Liu2025}. By retrieving evidence-based guidelines, clinical protocols, and documented practices, RAG systems minimize counterfactual mistakes. Second, RAG enables integration of evolving content and current research without continuous retraining \citep{Pyae2024}. For school-based SLP, this capability is critical as ASHA guidelines and state standards evolve continuously. Third, RAG facilitates domain-specific knowledge integration by retrieving information from specialized databases, enabling contextualized responses grounded in clinical guidelines and institutional protocols \citep{Amugongo2025}.

However, RAG architecture combined with engineered prompting strategies applied to school-based SLP case generation remains unexplored. Integration of ASHA clinical guidelines with school-based documentation exemplars through RAG, coupled with systematic prompt template development, represents a novel approach potentially addressing limitations of conversational AI while enabling scalable generation of diverse, clinically appropriate, structurally consistent materials. Systematic evaluation frameworks for RAG-based healthcare applications remain limited \citep{Amugongo2025}, with most studies failing to address ethical considerations or establish standardized quality metrics. Whether RAG-based approaches can successfully generate high-quality school-based SLP vignettes, IEP goals, and session notes suitable for educational practice, student assessment, and research remains an open empirical question.

\subsection{Study Purpose}

This proof-of-concept study establishes technical feasibility for an AI-powered system generating comprehensive school-based SLP simulation cases. The system integrates retrieval-augmented generation (RAG) with curated domain knowledge to address two fundamental challenges: (1) general-purpose LLMs lack school-based SLP expertise and produce clinically inappropriate outputs requiring extensive revision, and (2) effective prompting requires dual clinical-technical expertise that limits scalable adoption and produces inconsistent outputs across users.

We additionally investigate whether open-source models deployed locally can generate clinically appropriate cases, addressing practical concerns about accessibility, cost, and data privacy for resource-limited institutions. Comparing performance between premium commercial models (GPT-4o, Claude 3.5 Sonnet, Gemini 2.5 Pro) and open-source locally-deployed models (Llama 3.2, Qwen 2.5) provides preliminary benchmarks informing future validation studies and institutional model selection decisions.

This work demonstrates system capabilities and technical feasibility only. Extensive validation through expert review is required before any educational, research, or clinical implementation. The generated cases are not ready for educational use without rigorous validation. Rather, the proof-of-concept is a foundation for future validation studies examining whether RAG-augmented generation can serve educational practice, research applications, or assessment development in school-based SLP contexts.

%% file: 3-methods.tex
\section{Method}

This proof-of-concept system addresses two fundamental challenges in school-based speech-language pathology vignette generation: domain knowledge gaps in general-purpose large language models that produce clinically inappropriate outputs, and prompting inconsistency that limits scalable adoption. The methodological approach integrates retrieval-augmented generation with engineered prompt templates to enable consistent, evidence-based case generation without requiring specialized artificial intelligence training from end users.

\subsection{System Architecture} 

\begin{figure}[H]
    \centering
    \includegraphics[width=0.7\linewidth]{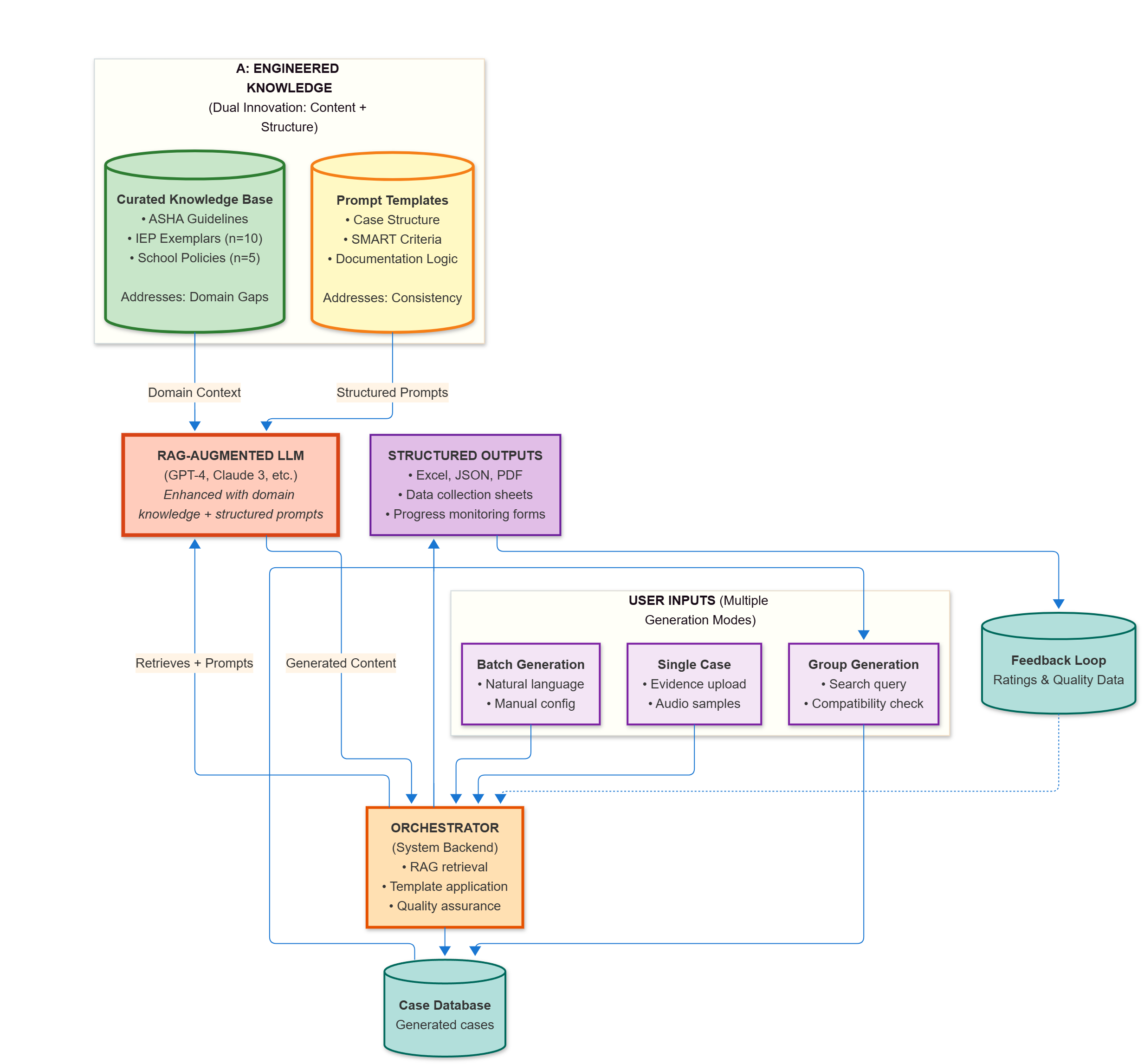}
    \caption{System Architecture}
    \label{fig:sys_arc}
\end{figure}

The system was developed using Claude Code \citep{Anthropic2025} through iterative prototyping of the RAG pipeline and prompt engineering architecture. The multi-component architecture implements three design principles: modularity supporting multiple large language model backends without model-specific retraining, evidence-grounding through RAG-based retrieval from curated authoritative sources, and structural consistency via engineered prompt templates encoding expert clinical knowledge (see Figure \ref{fig:sys_arc}).

System components include: (1) engineered knowledge foundation comprising a curated knowledge base and systematically designed prompt templates, (2) RAG-augmented large language model interface accessing commercial and open-source models through unified API, (3) orchestrator backend coordinating knowledge retrieval, prompt application, model inference, and output formatting, (4) case database providing persistent storage for case reuse and group session planning, (5) automatic speech recognition pipeline processing audio samples into de-identified transcripts with pattern detection and AI clinical analysis, and (6) feedback system collecting evaluations, checking grammatical accuracy, categorizing generation errors, and populating structured database for continuous system improvement.

\subsection{System Scope and Coverage}

The system supports coverage of 11 specific disorder types across 6 major categories, encompassing the full range of communication disorders addressed by school-based speech-language pathologists under IDEA eligibility categories \citep{IDEA2004}. Table \ref{tab:disorder_coverage} presents the complete taxonomy of supported disorder types with associated standardized assessments integrated into case generation. 

\begin{table}[ht!]
\centering
\small
\caption{Disorder Type Coverage and Assessment Specifications}
\label{tab:disorder_coverage}
\begin{tabularx}{\textwidth}{ l >{\RaggedRight\arraybackslash}X >{\RaggedRight\arraybackslash}X >{\RaggedRight\arraybackslash}X }
\toprule
\textbf{Disorder Category} & \textbf{Specific Disorder Type} & \textbf{Standardized Assessment(s)} & \textbf{Assessment Domain(s)} \\
\midrule
\multirow{3}{*}{Speech Sound Disorders} & Articulation Disorders & GFTA-3 (Goldman-Fristoe Test of Articulation-3) & Consonant production in words \\
 & Phonological Disorders & KLPA-3 (Khan-Lewis Phonological Analysis-3) & Phonological processes and error patterns \\
 & Speech Sound Disorder (general/mixed) & GFTA-3, KLPA-3 & Combined articulation and phonological assessment \\
\midrule
\multirow{3}{*}{Language Disorders} & Expressive Language Disorders & CELF-5 Expressive Language Index; EVT-3 (Expressive Vocabulary Test-3) & Sentence formulation, word structure, expressive vocabulary, word retrieval \\
 & Receptive Language Disorders & CELF-5 Receptive Language Index; PPVT-5 (Peabody Picture Vocabulary Test-5) & Sentence comprehension, semantic relationships, receptive vocabulary \\
 & Language Disorders (general/mixed) & CELF-5 Core Language Score & Overall receptive and expressive language performance \\
\midrule
\multirow{2}{*}{Social Communication} & Pragmatic Language Disorders & CASL-2 Pragmatic Language (Comprehensive Assessment of Spoken Language-2) & Pragmatic language skills and social language use \\
 & Social Communication Disorders & CASL-2 Pragmatic Language & Social communication abilities and pragmatic competence \\
\midrule
Fluency & Fluency Disorders & SSI-4 (Stuttering Severity Instrument-4) & Stuttering frequency, duration, and physical concomitants \\
\midrule
Motor Speech & Childhood Apraxia of Speech & VMPAC (Verbal Motor Production Assessment for Children) & Motor speech control and speech motor planning \\
\midrule
Voice & Voice Disorders & CAPE-V (Consensus Auditory-Perceptual Evaluation of Voice); PVOS (Pediatric Voice Outcome Survey) & Voice quality characteristics; voice-related quality of life \\
\bottomrule
\multicolumn{4}{l}{\small\textit{Note:} System scope limited to school-based speech, language, and communication impairments.} \\
\end{tabularx}
\end{table}
The system includes both pragmatic language disorders and social communication disorders as distinct categories, though both utilize CASL-2 Pragmatic Language assessment. This distinction reflects diagnostic variability in school-based practice where pragmatic deficits may occur with or without broader social communication impairments \citep{ASHA2024SCD}. Swallowing/feeding disorders were excluded as they require medical settings and instrumental assessments (MBSS, FEES) beyond standard school-based practice. Total coverage: 11 specific disorder types across 6 major categories.

\subsection{Knowledge Base Construction}

The knowledge base comprises 44 documents across four collections totaling 3,233 embedded chunks. Clinical practice guidelines (14 PDFs) sourced from ASHA Practice Portal cover speech sound disorders, language disorders, childhood apraxia of speech, fluency disorders, voice disorders, resonance disorders, augmentative and alternative communication, and cultural responsiveness \citep{ASHA2024}. Developmental milestone research (15 PDFs) includes peer-reviewed articles documenting normative trajectories for phonological development \citep{Porter2001, Preisser1988}, consonant acquisition \citep{Crowe2020}, vocabulary development \citep{Bornstein2004}, fluency development \citep{Ambrose1999}, and phonological process decline \citep{Smit1990}. IEP exemplars (10 documents) demonstrate goal-writing standards, measurable annual goals, and progress monitoring procedures \citep{Bateman2006, Yell2017}. School policy guidance (5 documents) addresses service delivery models, standards-based IEP requirements, and compliance standards \citep{IDEA2004}.

Documents underwent standardized preprocessing using LangChain framework \citep{Chase2022}. PDFs were loaded with PyPDFLoader using multithreading. Text segmentation employed RecursiveCharacterTextSplitter with 1,200-character chunks and 200-character overlap balancing semantic coherence with retrieval precision \citep{Zhao2023}. Each chunk received contextual metadata including source type, collection category, file identifiers, and date fields. Vector embeddings generated using OpenAI's text-embedding-3-small model \citep{OpenAI2024} produce 1,536-dimensional representations optimized for semantic similarity search. Embedded chunks were stored in ChromaDB \citep{Troynikov2023}, an open-source vector database supporting approximate nearest neighbor search with metadata filtering.

\subsection{Prompt Engineering}

Prompt templates implement established best practices for domain-specific text generation \citep{White2023, Zhou2023} through iterative refinement based on generation trials and expert review. The system employs a dual-prompt architecture with model-specific optimization: a comprehensive 493-line prompt for commercial premium models and a focused 281-line prompt for open-source models. This design reflects empirical findings during development that prompt complexity must match model instruction-following capacity. Initial testing with a unified comprehensive prompt across all models revealed that smaller open-source models (Llama 3.2, Qwen 2.5-7B) struggled with extended multi-step instructions, producing incomplete JSON structures with missing required fields despite perfect RAG retrieval. The focused free-model prompt maintains core clinical requirements while reducing cognitive load through simplified language and condensed formatting specifications, enabling reliable structured output from resource-constrained models.

Templates incorporate four components: (1) context integration injecting retrieved knowledge base chunks into prompt prefix, (2) structured output specifications enforcing JSON schema compliance with required fields for demographics, background information, assessment results, IEP goals, and session notes \citep{Ouyang2022}, (3) clinical constraints ensuring developmental appropriateness, disorder-goal alignment, exclusive goal targeting, and baseline data realism, and (4) cultural responsiveness requirements mandating authentic family structures, linguistic environments, and culturally relevant intervention activities addressing documented disparities \citep{Artiles2010, Sullivan2013}.

\subsection{Generation Modes}

The system implements three generation modes. Single case generation produces complete case files with user-specified grade level and disorder type, retrieving ten most relevant knowledge base chunks, constructing contextualized prompts, and generating dual-format output (structured JSON and Excel spreadsheet). Generated cases include demographics with culturally appropriate pseudonyms, comprehensive background information, standardized assessment results, 2-3 measurable IEP goals with baseline data, and three therapy session notes with objective performance data.

Multiple case batch generation enables dataset creation through three input methods: manual configuration via interactive forms, natural language parsing extracting parameters from free-text requests, or CSV/Excel upload with pre-specified student rosters. Algorithmic pseudonym generation ensures culturally diverse naming with appropriate demographic consistency \citep{Caliskan2017}. Diversity control parameters specify distributions across disorder types, grade levels, severity ranges, and cultural backgrounds.

Group session generation addresses small-group service delivery \citep{Cirrin2010}. The system searches existing cases for compatible students based on maximum two-year grade difference, disorder combinations conducive to shared activities, and severity matching. Insufficient matches trigger new case generation. Secondary LLM calls synthesize collaborative session plans integrating individual goals into shared activities with differentiated targets.

\subsection{Pipelines}
\subsubsection{RAG Pipeline}

The RAG pipeline implements standard retrieve-then-generate architecture \citep{Gao2024}. For each request, structured queries combining disorder type, grade level, and population characteristics undergo embedding using text-embedding-3-small model ensuring semantic alignment with stored chunks. ChromaDB performs approximate nearest neighbor search computing cosine similarity between query and chunk embeddings, returning k=10 most similar chunks with metadata and source text. Retrieved chunks undergo concatenation with metadata headers and injection into prompt template context fields. Large language models generate content conditioned on both retrieved context providing domain knowledge and structured prompts providing output schema and clinical constraints, mitigating hallucination in specialized domains \citep{Ji2023}.

\subsubsection{Large Language Model Configuration}

The system supports cloud-based commercial models and locally deployed open-source models. Premium commercial models include GPT-4o \citep{OpenAI2024}, Gemini 2.5 Pro \citep{GoogleDeepMind2024}, Claude 3 Opus and Claude 3.5 Sonnet \citep{Anthropic2024}, accessed via API with temperature=0.7 balancing creativity with consistency. Open-source models include Llama 3.2 \citep{MetaAI2024}, Qwen 2.5 \citep{AlibabaCloud2024}, and DeepSeek R1 \citep{DeepSeekAI2024}, deployed locally using Ollama \citep{OllamaInc2024} ensuring data privacy and offline operation. Dropdown interface enables model selection for direct performance comparison using identical prompts and retrieved context.

\subsubsection{ASR and Clinical Analysis Pipeline}

The automatic speech recognition pipeline processes audio samples (WAV, MP3, M4A) for educational transcription and diagnostic reasoning exercises. Whisper base model performs transcription with time-stamped utterance boundaries. De-identification employs regex patterns detecting and replacing names, phone numbers, email addresses, street addresses, and dates with generic placeholders while preserving clinical relevance.

Pattern detection analyzes phonological characteristics identifying sound repetitions, syllable repetitions, prolongations, and blocks for fluency assessment. Articulation error detection identifies phoneme substitutions, omissions, and distortions. Language pattern analysis computes mean length of utterance approximations, average word and sentence length, and identifies morphological errors. Detected patterns undergo AI-powered clinical analysis using local or cloud-based large language models generating diagnostic hypotheses, severity ratings, estimated age ranges, recommended IEP goals, clinical observations, and evidence-based intervention recommendations. Analysis output models clinical reasoning processes supporting student skill development through comparative feedback.

\subsubsection{Feedback System}

The feedback collection mechanism provides structured evaluation forms for expert reviewers rating clinical accuracy, documentation quality, educational utility, and cultural appropriateness using five-point Likert scales with open-ended text fields. Submitted evaluations populate relational database with foreign keys linking feedback to specific cases, timestamps enabling temporal analysis, and reviewer identifiers supporting inter-rater reliability calculation.

Automated grammar checking analyzes generated text using rule-based parsers and neural language models detecting syntactic violations, misspellings, punctuation inconsistencies, and unnatural phrasing. Detected errors receive categorical classification by type and severity flags. Generation error taxonomy includes developmental inappropriateness, disorder-goal misalignment, internal inconsistency, documentation standard violations, and cultural insensitivity. Aggregated error analysis generates system-level quality reports identifying prevalent patterns, systematic quality variations by disorder type, model-specific performance differences, and temporal trends informing targeted prompt refinement, knowledge base augmentation, and model configuration optimization.

\subsection{Implementation}

The system employs Python with Gradio framework \citep{Abid2019} providing web-based interface with tabbed navigation, dropdown menus, text input boxes, and output components displaying JSON with syntax highlighting and interactive Excel tables. Modular architecture separates UI components (app/ui\_*.py), generation logic (app/generation.py), utility functions (app/utils.py), and configuration management (app/config.py). Dependencies managed via Python virtual environment with pinned versions ensuring reproducible installations. Two deployment contexts support local execution and cloud deployment on Hugging Face Spaces.

\subsection{Preliminary Validation}

To evaluate system performance across the intended design space and compare generation quality between commercial and open-source models, complete cases were generated for seven test scenarios spanning the disorder taxonomy (Table \ref{tab:test_cases_compact}): speech sound disorders (2nd grade articulation), mixed receptive-expressive language disorders (4th grade), pragmatic language disorders (6th grade), fluency disorders (9th grade), voice disorders (Pre-K), combined phonological/expressive language disorders (Kindergarten), and fluency with pragmatics (10th grade). Test cases were systematically selected to represent diverse grade levels (Pre-K through high school), and disorder complexity (single vs. co-occurring disorders).
\begin{table}[ht!]
\centering
\small
\caption{Test Case Distribution Across Disorder Types and Grade Levels}
\label{tab:test_cases_compact}
\begin{tabularx}{\textwidth}{ >{\RaggedRight\arraybackslash}X c c }
\toprule
\textbf{Disorder Type(s)} & \textbf{Grade Level} & \textbf{Cases per Model} \\
\midrule
Articulation Disorders & 2nd Grade & 1 \\
Mixed Receptive-Expressive Language & 4th Grade & 1 \\
Pragmatic Language (single disorder) & 6th Grade & 1 \\
Fluency Disorders & 9th Grade & 1 \\
Voice Disorders & Pre-K & 1 \\
Combined Phonological-Expressive Language & Kindergarten & 1 \\
Combined Pragmatic-Expressive Language & 1st Grade & 1 \\
\midrule
\textbf{Total per Model} & \textbf{Pre-K to 9th} & \textbf{7} \\
\bottomrule
\multicolumn{3}{l}{\footnotesize \textit{Note:} Five models evaluated (GPT-4o, Claude 3.5 Sonnet, Gemini 2.5 Pro, Llama 3.2, Qwen 2.5-7B),} \\
\multicolumn{3}{l}{\footnotesize producing 35 total validation cases. Test cases systematically represent single-disorder presentations,} \\
\multicolumn{3}{l}{\footnotesize mixed presentations, and combined disorders across diverse grade levels.} \\
\end{tabularx}
\end{table}

Cases were generated using five large language models representing different accessibility contexts: three premium commercial models (GPT-4o via OpenAI API, Claude 3.5 Sonnet via Anthropic API, Gemini 2.5 Pro via Google API) and cloud-based processing, and two open-source models (Llama 3.2, Qwen 2.5-7B) deployed locally using Ollama framework enabling offline operation and institutional data privacy. Each of the seven test scenarios was generated once per model, producing 35 total validation cases. All cases utilized identical RAG retrieval parameters (k=10 most relevant knowledge base chunks), prompt templates, and generation settings (temperature=0.7) to isolate model-specific performance differences from system configuration effects.

\subsubsection{Automated Quality Assessment} 

To establish initial quality baselines prior to expert clinical review, generated validation cases underwent automated computational evaluation using Claude Sonnet 4.5 \citep{Anthropic2025} as an assessment tool. This approach provides preliminary quality indicators through systematic, reproducible scoring but does not substitute for expert SLP validation of clinical appropriateness.

Each case was assessed across four quality dimensions using a structured evaluation rubric developed for this proof-of-concept study (\ref{tab:quality_assess_rubric}). The rubric was informed by established SMART goal criteria (Specific, Measurable, Achievable, Relevant, Time-bound) widely used in IEP development \citep{Hedin2018}, with additional consideration of emerging frameworks emphasizing specificity and measurability in SLP goal setting \citep{Hamilton2025}. The rubric evaluated: (1) structural completeness of required case components, (2) internal consistency between background information, assessment results, and intervention goals, (3) clinical appropriateness, including developmental expectations and disorder-specific content, and (4) documentation quality, including adherence to SMART criteria for annual goals and objective data collection in session notes.

Scoring employed a 5-point scale where 1=major deficiencies requiring complete revision, 3=acceptable with notable limitations, and 5=high quality meeting professional standards. The four dimensions were:
\begin{table}[ht!]  
    \centering
    \caption{Quality Assessment Rubric}
    \label{tab:quality_assess_rubric}
    \begin{tabularx}{\textwidth}{ >{\RaggedRight}p{4.5cm} >{\RaggedRight\arraybackslash}X }
        \toprule
        \textbf{Criterion} & \textbf{Description} \\
        \midrule
        Structural Completeness (1-5) & Presence and completeness of all required data fields including student demographics, background information with medical history, parent, and teacher concerns, standardized assessment results with appropriate instruments, annual IEP goals, and therapy session notes. Scoring reflected proportion of required fields present and adequately populated. \\
        \addlinespace
        Internal Consistency (1-5) & Coherence and logical alignment between case components. Evaluation examined whether background information supports the identified disorders, assessment results align with stated concerns and disorder presentations, annual goals appropriately target identified deficits and assessment findings, and session notes address goals with intervention activities matched to disorder types and developmental levels. \\
        \addlinespace
        Clinical Appropriateness (1-5) & Developmental appropriateness of expectations and activities for specified grade level, realism of disorder presentations and symptom patterns, appropriate assessment instrument selection for disorder types and ages, realistic severity characterizations aligned with score distributions, and sufficient background detail length (minimum 300 characters capturing meaningful clinical context). \\
        \addlinespace
        IEP Goal \& Session Notes Quality (1-5) & Adherence to professional documentation standards include SMART criteria for annual goals (Specific behavior, Measurable criterion with percentage or trial counts, Achievable within one year, Relevant to educational access, Time-bound with "before or by next annual ARD"), appropriate context and timeframe specifications, and measurable data in session notes with objective trial counts or percentage accuracy. \\
        \bottomrule
    \end{tabularx}    
\end{table}

Automated scoring was performed using structured prompts directing the evaluation model to assess each dimension independently, identify specific issues, assign scores, and provide justification. This computational approach enables rapid, consistent evaluation across large case sets but cannot detect subtle clinical errors requiring domain expertise such as inappropriate developmental expectations, internally contradictory assessment interpretations, or culturally insensitive content.

%% file: 4-results.tex
\section{Results}
This section presents system implementation outcomes and demonstrates the system's capability to generate complete school-based SLP simulation cases across diverse clinical contexts.

\subsection{System Implementation}

The system was successfully implemented with support for multiple large language models including premium commercial models (GPT-4o, Claude 3.5 Sonnet, Gemini 2.5 Pro) and open-source locally-deployed models (Llama 3.2, Qwen 2.5-7B), enabling flexibility for institutions with different budget constraints, privacy requirements, and infrastructure capabilities. The modular architecture allows straightforward integration of additional models as they become available without requiring system redesign or knowledge base modification.

All three intended generation modes were implemented and operationally tested: (1) single case generation creates complete case files on demand with user-specified parameters, (2) batch generation produces up to 100 cases simultaneously with controlled diversity across disorder types and demographics, and (3) group session planning generates matched student profiles for small-group therapy scenarios with compatibility considerations based on grade proximity and disorder combinations.

The system produces output in multiple formats aligned with educational workflows. Excel spreadsheets provide editable case files with structured data accessible to clinical faculty for customization. JSON format enables machine-readable output for integration with learning management systems or automated assessment platforms. PDF format produces formatted documents suitable for printing and distribution in paper-based simulation exercises.

\subsection{Technical Performance Across Case Generation}

Across 35 systematic test generations spanning all disorder categories (Table \ref{tab:disorder_coverage}), grade levels, and model types, the system demonstrated robust operational performance. Generation success rate was 100\%, with all cases producing structurally complete output files containing all required components: student demographics with culturally appropriate pseudonyms, comprehensive background information including medical history and teacher concerns, standardized assessment results with appropriate instruments, 2-3 measurable annual IEP goals formatted according to SMART criteria, and three longitudinal therapy session notes with objective performance data. No cases required regeneration due to structural incompleteness, missing sections, or technical failures.

Generation latency varied by model type and infrastructure, with commercial API-based models generally responding faster than locally-deployed open-source models on standard hardware. RAG retrieval latency remained consistent across all models, confirming that knowledge base retrieval does not constitute a performance bottleneck. 

The user interface implements both conversational and structured manual input modes. The conversational interface supports novice users through guided prompts and clarifying questions, enabling case generation through natural language requests without technical knowledge of system parameters. The structured interface enables experienced users to manually select specific parameters including disorder type, grade level, and demographic characteristics without conversational interaction, facilitating rapid batch generation for research protocols or examination development.

\subsection{Performance Across Models.} 
Table \ref{tab:rating} presents automated quality scores averaged across the seven test cases per model, revealing performance differences between commercial and open-source implementations.
\begin{table}[htp] 
    \centering
    \small
    \caption{Automated Quality Ratings}
    \label{tab:rating}
    \begin{tabularx}{\textwidth}{ l l *{5}{>{\centering\arraybackslash}X} }
        \toprule
        \textbf{Model} & \textbf{Type} & \textbf{Struct. Complete.} & \textbf{Internal Consist.} & \textbf{Clinical Approp.} & \textbf{IEP/Note Quality} & \textbf{Overall Average} \\
        \midrule
        GPT-4o & Commercial & 5.00 & 3.71 & 4.29 & 5.00 & \textbf{4.50} \\
        Claude 3.5 Sonnet & Commercial & 5.00 & 3.71 & 4.57 & 4.57 & \textbf{4.46} \\
        Gemini 2.5 Pro & Commercial & 5.00 & 3.57 & 4.29 & 4.71 & \textbf{4.39} \\
        Llama 3.2 & Open-Source & 5.00 & 2.86 & 4.00 & 4.86 & \textbf{4.18} \\
        Qwen 2.5-7B & Open-Source & 5.00 & 3.29 & 4.43 & 4.29 & \textbf{4.25} \\
        \bottomrule
        \addlinespace[2pt] 
        \multicolumn{7}{l}{\footnotesize Note: Scores range 1-5 where 5 indicates highest quality. Averages computed across 7 test cases per model.} \\
    \end{tabularx}
\end{table}
Premium commercial models achieved marginally higher average scores (range: 4.39-4.50) compared to open-source models (range: 4.18-4.25), a difference of approximately 0.2-0.3 points on the 5-point scale. All five models achieved perfect structural completeness scores (5.00), confirming that the RAG architecture and prompt engineering successfully constrain output format across diverse model capabilities. Performance variation emerged primarily in internal consistency, where commercial models averaged 3.66 compared to open-source models averaging 3.08, and clinical appropriateness, where commercial models averaged 4.38 compared to open-source models averaging 4.22.

\subsubsection{Common Quality Patterns} 
The most frequent issue across all models was session notes failing to explicitly reference IEP goal numbers (observed in 23 of 35 cases, 66\%), despite intervention activities appropriately targeting goal domains. This documentation gap represents a formatting issue rather than clinical misalignment and could be addressed through prompt template refinement specifying explicit goal numbering requirements. The second most common issue was background information occasionally omitting mention of one co-occurring disorder when multiple disorders were specified (observed in 6 cases, 17\%), suggesting the need for enhanced prompts enforcing comprehensive disorder coverage in background narratives.
\begin{figure} [H]
    \centering
    \includegraphics[width=0.9\linewidth]{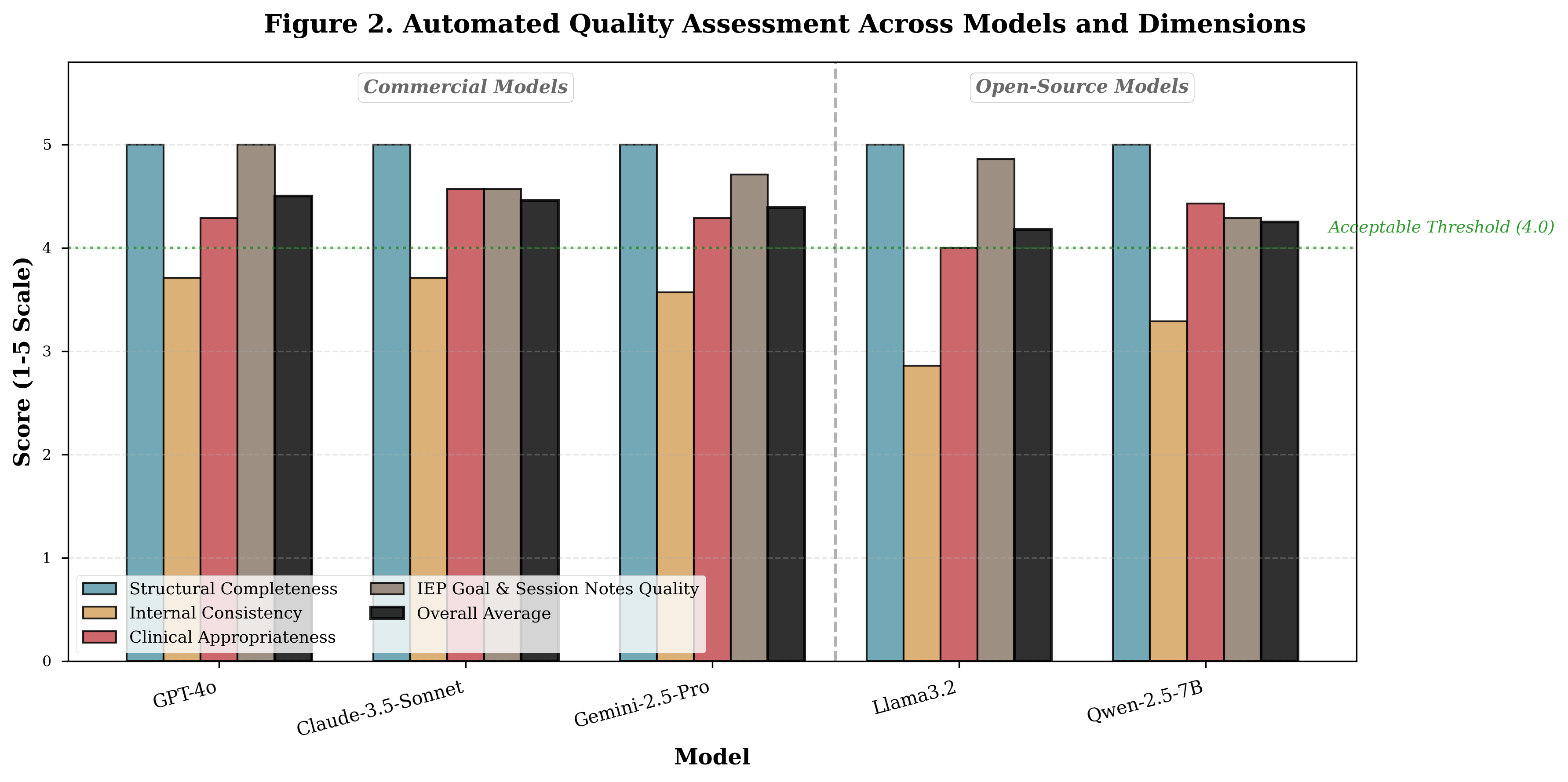} \caption{Model Performance Comparison}   
    \label{fig:performancecompmodel}
\end{figure}
Open-source models demonstrated specific patterns: Llama 3.2 generated backgrounds occasionally below the 300-character minimum (3 of 7 cases), and both open-source models occasionally produced IEP goals lacking specific measurable criteria or timeframe context (5 of 14 cases combined). Commercial models showed greater consistency in SMART goal formatting and appropriate background detail length. However, open-source model performance remained within acceptable ranges (all scores $\geq$4.00 except internal consistency), suggesting that with identical RAG retrieval and prompt templates, smaller open-source models can generate structurally appropriate, clinically reasonable cases suitable for preliminary educational applications or pilot testing (Figure \ref{fig:performancecompmodel}). 

\subsubsection{Performance Across Disorder Types}
\begin{figure} [H]
    \centering
    \includegraphics[width=0.8\linewidth]{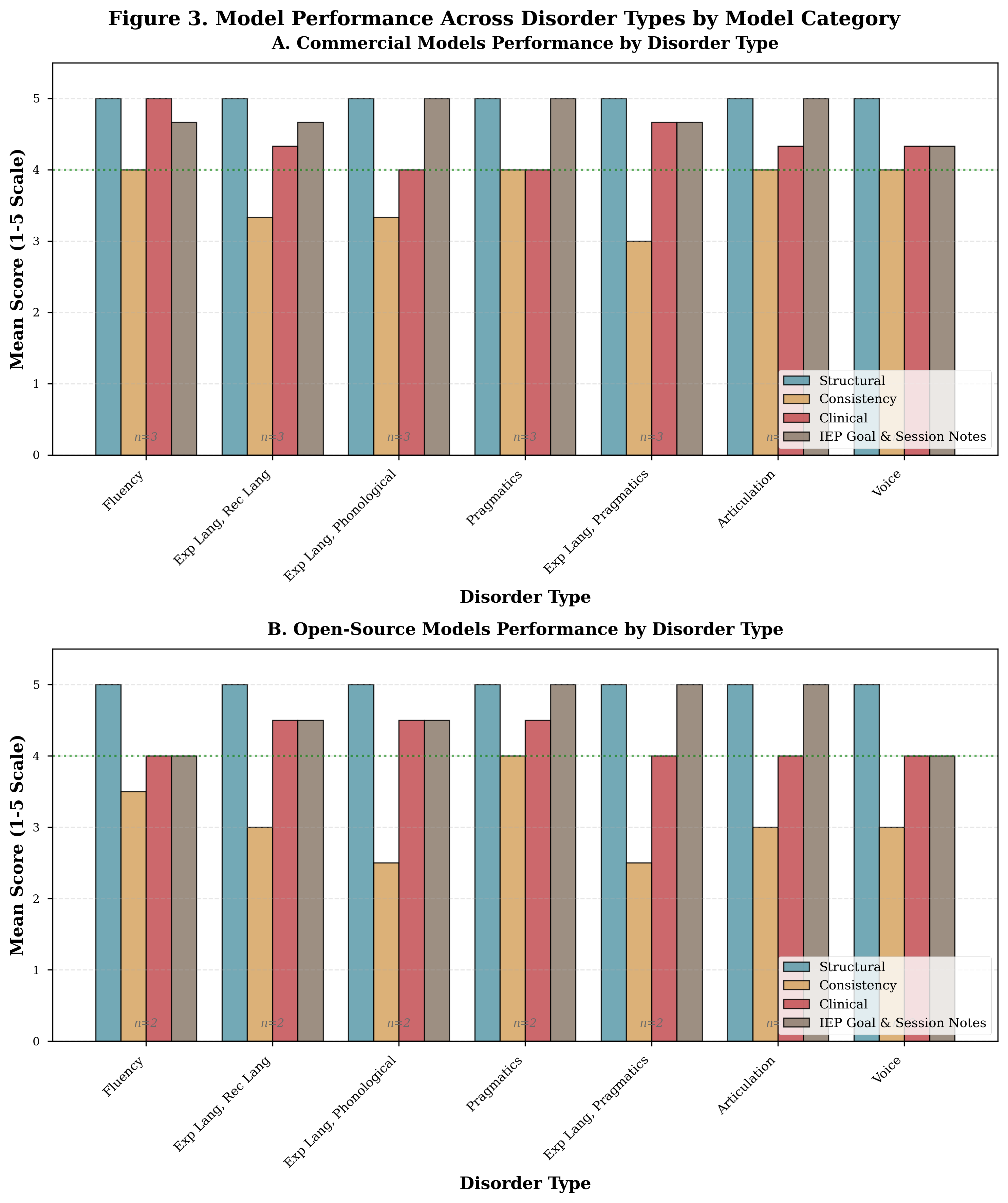}
    \caption{Disorder Performance by Category}
     \label{disorder_perf}
\end{figure}

Performance patterns are varied systematically across disorder types (Figure \ref{disorder_perf}). Articulation disorders produced relatively balanced scores across all dimensions (4.0-5.0), representing one of the most successfully generated disorder types. For fluency disorders, both model categories demonstrated strong structural completeness (5.0) but struggled with internal consistency (commercial: 4.0; open-source: 3.5), suggesting difficulty integrating stuttering assessment data with intervention planning. Expressive and receptive language cases showed the poorest internal consistency scores across both categories (commercial: 3.3; open-source: 3.0), likely reflecting the complexity of coordinating multiple language domains (syntax, morphology, semantics) within a cohesive goal setting and activity plan. Expressive language paired with phonological disorders yielded moderate consistency scores (commercial: 3.3; open-source: 2.5), with open-source models particularly challenged by the dual-disorder complexity. Pragmatics cases demonstrated stronger performance across dimensions (consistency: commercial 4.0, open-source 4.0), possibly due to more straightforward social communication goal structures. Cases combining expressive language with pragmatics showed similar patterns (consistency: commercial 3.0, open-source 2.5), maintaining the trend of reduced consistency in multi-domain cases. Voice disorders similarly showed strong overall performance (commercial: 4.0-5.0; open-source: 3.0-5.0), though open-source models exhibited lower consistency (3.0) compared to their commercial counterparts. Commercial models consistently outperformed open-source models by 0.5-1.0 points on internal consistency across all disorder types, while maintaining comparable performance on other dimensions.

%% file: 5-discussion.tex
\section{Discussion}

This proof-of-concept study demonstrates the technical feasibility of integrating RAG with engineered prompt templates to generate school-based speech-language pathology simulation cases. The system successfully addressed two fundamental challenges in AI-assisted case generation: domain knowledge gaps inherent in general-purpose large language models, and the dual clinical-technical expertise barrier that limits scalable adoption of AI tools in specialized clinical domains. By embedding expert knowledge into reusable system architecture through curated knowledge bases and validated prompt templates, the system enables consistent generation of structurally complete, clinically grounded cases across diverse disorder types and grade levels without requiring end users to develop sophisticated AI interaction skills.

\subsection{Technical Performance}

The system achieved 100\% structural completeness across all 35 validation cases, demonstrating that RAG architecture combined with engineered prompts successfully constrains output format regardless of underlying model capabilities. This structural consistency represents a significant advancement over ad hoc conversational AI approaches, where output quality varies substantially based on individual user prompting skills \citep{Kim2025}. Similar findings have been reported in medical education, where structured prompt engineering improved consistency of AI-generated clinical scenarios compared to unstructured approaches \citep{Bakkum2024}. Generated cases consistently included all required components: demographics with culturally appropriate pseudonyms, comprehensive background information integrating medical history and educational concerns, standardized assessment results with disorder-appropriate instruments, measurable annual IEP goals formatted according to SMART criteria, and longitudinal therapy session notes with objective performance data.

The system's multi-model architecture provides institutional flexibility across different resource contexts. Premium commercial models achieved marginally higher automated quality scores (range: 4.39-4.50) compared to open-source locally-deployed models (range: 4.18-4.25), a difference of approximately 0.2-0.3 points on the 5-point scale. This relatively modest performance gap aligns with recent comparative evaluations showing that knowledge retrieval mechanisms can partially mitigate performance differences between commercial and open-source language models in domain-specific tasks \citep{Gao2024, Chen2025}. With identical RAG retrieval mechanisms and model-appropriate prompt templates, open-source models can generate structurally appropriate and clinically reasonable cases suitable for preliminary educational applications, addressing practical concerns about cost, data privacy, and institutional accessibility for resource-limited settings.

The dual-prompt architecture emerged from empirical testing during system development. Initial attempts to use a single comprehensive prompt (493 lines) across all models resulted in structural failures for open-source models, with generated cases missing core fields such as background information, annual goals, and assessment results. This finding reveals an important constraint in prompt engineering: instruction complexity must align with model instruction-following capacity. While comprehensive prompts with extensive validation checks and detailed examples benefit larger commercial models, they overwhelm smaller open-source models (3-7B parameters), causing incomplete structured output despite successful knowledge retrieval. The focused 281-line prompt for open-source models maintains essential clinical constraints while reducing cognitive load, achieving reliable JSON schema compliance. This architectural decision prioritizes correctness over simplicity, accepting maintenance complexity to ensure structural validity across diverse computational resources.

\subsection{Quality Patterns Across Disorder Types }

Automated quality assessment revealed systematic patterns that inform future refinement. Internal consistency scores showed the greatest variability across models and disorder types, with commercial models averaging 3.66 compared to open-source models at 3.08. Language disorders, particularly those involving multiple domains, consistently yielded lower consistency scores across both model categories, likely reflecting the inherent complexity of coordinating syntax, morphology, semantics, and phonology within cohesive goal structures. This pattern parallels findings in natural language generation research, where multi-constraint optimization tasks consistently demonstrate greater difficulty than single-domain generation \citep{Ji2023, Kumar2021}. Articulation and pragmatic language disorders demonstrated stronger consistency, possibly due to more straightforward goal structures and fewer interacting domains.

\subsection{Implications Across Multiple Domains}

\subsubsection{Clinical Practice Implications}

The system's demonstrated capacity to generate clinically grounded cases with consistent structure addresses a critical need in school-based speech-language pathology training, where access to diverse clinical presentations is often limited by geographic constraints, caseload composition, and privacy requirements \citep{Dudding2018}. Generated cases provide exposure to disorder presentations that trainees may not encounter during clinical placements, including low-incidence conditions, complex co-occurring disorders, and diverse cultural-linguistic backgrounds. The consistency of case structure, with standardized assessment results, SMART-formatted goals, and longitudinal session notes, models professional documentation practices that novice clinicians often struggle to master \citep{Hamilton2025}.

The group session generation functionality represents real-world clinical decision-making in school settings, where clinicians must balance individualized intervention with scheduling constraints and shared treatment activities. By generating compatible case groupings, the system provides practice opportunities for complex clinical reasoning about appropriate grouping criteria, shared intervention activities, and simultaneous progress monitoring for multiple students. This mirrors authentic clinical workflows where practitioners must consider not only individual student needs but also pragmatic service delivery factors \citep{Cirrin2010}.

\subsubsection{Educational Implementation Implications}

For educational programs, the system offers potential to address several pedagogical challenges. The batch generation capability enables the creation of standardized case sets for assessment purposes, supporting program-level evaluation of student competencies across consistent materials while varying disorder presentations and complexity levels. This addresses recurring concerns about assessment validity when human-authored cases may inadvertently vary in difficulty or completeness \citep{Peabody2000}. The system's multi-modal interface, supporting both conversational and structured input, accommodates different instructor preferences and pedagogical approaches, from exploratory learning activities to structured assessment scenarios.

The ability to rapidly generate cases tailored to specific learning objectives enables curriculum integration at multiple levels. Instructors can align case characteristics with progressive skill development, introducing simpler cases early in programs and increasing complexity as students develop clinical reasoning capabilities. The system facilitates deliberate practice with immediate case availability, reducing instructor preparation burden while maintaining pedagogical control over learning objectives and complexity progression \citep{Sridharan2024, Duvivier2011}. However, educational effectiveness depends critically on thoughtful integration within broader curriculum design, not simply case availability.

\subsubsection{Research Applications and Opportunities}

The system creates novel research opportunities in clinical education. Large-scale generation capabilities enable psychometric studies requiring substantial case sets with controlled variations in specific parameters while holding other factors constant. Researchers can systematically investigate how case complexity, disorder type, cultural background, or documentation quality influence student diagnostic accuracy or clinical reasoning processes. This controlled variation is challenging to achieve with human-authored cases, where confounding factors are difficult to isolate.

\subsubsection{Technology Development Contributions}

From a technical perspective, this work demonstrates that RAG with engineered prompts can address domain knowledge limitations in general-purpose language models for specialized applications. The relatively modest performance difference between commercial and open-source models when using identical RAG infrastructure suggests that architecture-neutral approaches to knowledge integration may democratize AI capabilities across institutions with varying resource constraints. This has implications beyond clinical education for any specialized domain requiring consistent, structured output grounded in professional standards.

The prompt engineering approach, embedding domain expertise into reusable templates rather than requiring per-query expertise, represents a model for reducing the technical barrier to AI adoption in specialized fields. By separating knowledge curation from daily system use, the architecture enables domain experts to contribute to system development without requiring programming skills, while allowing non-expert users to generate appropriate outputs without mastering prompt engineering \citep{White2023, Zhou2023}. This separation of concerns may facilitate AI integration in domains where technical and domain expertise rarely overlap.

\subsection{Study Limitations and Validation Needs}

This proof-of-concept demonstrates technical feasibility but has important limitations requiring acknowledgment before broader implementation. The system has not undergone expert clinical validation by experienced school-based SLPs. While automated quality scores provide preliminary benchmarks informed by established SMART criteria and IEP development frameworks, they cannot substitute for rigorous expert review assessing clinical realism, developmental appropriateness, intervention effectiveness, and cultural sensitivity. Expert validation remains the gold standard in simulation-based education \citep{Nakamura2024, Cowie2021}, and computational metrics alone cannot detect subtle clinical errors, developmentally inappropriate expectations, culturally insensitive content, or inappropriate evidence-based practice applications that experienced clinicians would identify.

The limited scale of systematic testing (35 validation cases across 7 scenarios and 5 models) provides preliminary performance characterization but insufficient evidence for broad generalizations. Comprehensive validation would require generating and evaluating hundreds of cases across all supported disorder combinations, grade levels, severity ranges, and cultural backgrounds to identify systematic quality issues or underperforming domains. Inter-rater reliability studies with multiple expert reviewers would establish consistency of quality judgments and identify areas requiring consensus guidelines. Additionally, the knowledge base, while substantially more comprehensive than general-purpose language model training, remains incomplete and potentially biased toward conditions well-represented in published literature. Rare disorder presentations, emerging intervention research, and culturally specific practice patterns may be underrepresented.

The system has not been tested with end users in authentic educational contexts. Usability studies are needed to determine whether the interface effectively supports intended workflows, whether generated outputs require substantial manual editing, and whether conversational and structured input modes meet user needs. More critically, student pilot studies must investigate whether practice with AI-generated cases improves clinical reasoning skills, diagnostic accuracy, or goal-writing quality compared to traditional instructional approaches. Educational effectiveness cannot be assumed based on technical performance alone \citep{Dudding2018, Sridharan2024}.

\subsection{Future Directions}

Future research should prioritize systematic expert validation employing multiple reviewers with diverse clinical backgrounds and established inter-rater reliability protocols. Comparative studies investigating whether expert-reviewed AI-generated cases achieve psychometric properties comparable to human-authored cases would establish their utility for assessment purposes. Randomized controlled trials comparing learning outcomes between students practicing with AI-generated versus traditional cases would provide critical evidence about educational effectiveness, examining diagnostic accuracy, clinical reasoning quality, and goal-writing skills across different instructional sequences.

Technical development should explore multi-agent architectures where specialized models generate different case components with separate consistency-checking agents, hybrid human-AI workflows leveraging complementary strengths, and integration with learning management systems or clinical documentation platforms. Research investigating optimal knowledge base composition, maintenance strategies, and retrieval mechanisms would inform system refinement, while domain-specific fine-tuning approaches might complement or provide alternatives to RAG. Ethical and sociocultural research employing participatory design approaches, bias detection systems, and implementation frameworks would ensure responsible AI adoption in clinical education contexts. These directions would advance both theoretical understanding and practical application of AI-augmented case generation for health professions education.

%% file: 6-conclusion.tex
\section{Conclusion}

This proof-of-concept establishes technical feasibility of RAG-augmented generation for school-based speech-language pathology simulation cases, demonstrating that integration of curated knowledge bases with engineered prompt templates enables the generation of structurally complete, clinically grounded vignettes without requiring specialized AI expertise from end users. The system successfully generates comprehensive case files spanning diverse disorder types, grade levels, and demographic characteristics, with performance differences between premium commercial and open-source models remaining relatively modest when provided with identical knowledge bases and prompt architectures.

However, extensive validation through expert clinical review, psychometric evaluation, and educational effectiveness studies remains essential before any implementation in educational practice, student assessment, or research applications. This work provides a foundation for systematic investigation of whether and under what conditions RAG-augmented generation can appropriately supplement traditional approaches to simulation-based learning in speech-language pathology education.

%% file: 8-appendices.tex
\clearpage
\appendix

\section{Prompt Templates}
\label{app:prompts}

This appendix provides excerpts from the actual engineered prompt templates used in the RAG-augmented case generation system. These prompts demonstrate how expert clinical knowledge and evidence-based constraints are encoded to ensure consistent, clinically appropriate output across different large language models.

\subsection{Premium Model Prompt (Complete System Prompt)}

The following prompt template is used for commercial premium models (GPT-4o, Claude 3.5 Sonnet, Gemini 2.5 Pro). This represents the complete system prompt with placeholders for dynamic RAG-retrieved context and user-specified parameters.

\begin{small}
\begin{verbatim}
You are an expert school-based speech-language pathologist creating
comprehensive, realistic case files for clinical education. Your expertise
encompasses assessment, diagnosis, IEP development, intervention planning,
and progress documentation across all communication disorder types in
school-age populations (Pre-K through 12th grade).

=== TASK ===
Generate a complete student case file in valid JSON format with the following
components:
1. Student demographics (name, age, grade, gender, background)
2. Assessment results from standardized tests
3. 2-4 measurable annual IEP goals in SMART format
4. 3 detailed therapy session notes showing progress over time

=== CONSTRAINTS AND REQUIREMENTS ===

DISORDER FOCUS:
{disorders}

GRADE LEVEL:
{grade}

POPULATION-SPECIFIC REQUIREMENTS:
{population_spec}

CULTURAL DIVERSITY:
Generate a culturally appropriate name reflecting diverse backgrounds
(Hispanic/Latino, African American, Asian American, Middle Eastern,
Caucasian, Native American, or multiracial). Ensure all aspects of the
case reflect cultural and linguistic considerations as appropriate.

AGE-GRADE ALIGNMENT:
Ensure the student's age is developmentally appropriate for the specified
grade level:
- Pre-K: 4-5 years
- Kindergarten: 5-6 years
- 1st Grade: 6-7 years
- 2nd Grade: 7-8 years
[continues through 12th grade]

=== RETRIEVED KNOWLEDGE BASE CONTEXT ===

The following passages have been retrieved from authoritative clinical
sources to inform your case generation. Use this context to ensure clinical
accuracy, evidence-based assessment selection, developmentally appropriate
goals, and realistic intervention strategies.

{context}

=== ASSESSMENT REQUIREMENTS ===

1. Select standardized assessments appropriate for the disorder type and
   grade level from evidence-based options (e.g., GFTA-3, CAAP-2 for
   articulation/phonology; CELF-5, OWLS-II for language; SSI-4 for
   fluency; CASL-2 for pragmatics).

2. Provide realistic standard scores:
   - Scores should reflect "moderate" severity typically qualifying for
     school-based services (standard scores 70-85, roughly 2nd-16th
     percentile range)
   - Include percentile ranks for interpretability
   - Ensure scores are consistent with described functional impact

3. Assessment domains should directly correspond to the disorder type
   specified. For example:
   - Articulation Disorders: phoneme-level production accuracy,
     intelligibility ratings
   - Language Disorders: syntax, morphology, vocabulary, comprehension
   - Fluency Disorders: stuttering frequency, severity rating,
     secondary behaviors
   - Pragmatic Disorders: social communication skills across contexts

=== IEP ANNUAL GOALS REQUIREMENTS ===

Generate 2-4 annual goals following SMART criteria (Specific, Measurable,
Achievable, Relevant, Time-bound):

STRUCTURE TEMPLATE:
"Before or by the next annual ARD, [STUDENT NAME] will [TARGET SKILL/
BEHAVIOR] given [CONDITION/SUPPORT LEVEL] in [ACCURACY CRITERION] as
measured by [DATA COLLECTION METHOD]."

REQUIREMENTS FOR EACH GOAL:
1. SPECIFIC: Precisely define the target skill with observable,
   operational terms
2. MEASURABLE: Include quantifiable success criteria (percentage accuracy,
   trial counts, frequency counts, duration measures)
   - Common criteria: "80% accuracy," "4 out of 5 trials," "in 8/10
     opportunities"
3. ACHIEVABLE: Appropriate for one-year timeline given student's current
   baseline performance
4. RELEVANT: Directly addresses the specified disorder and impacts
   educational/social participation
5. TIME-BOUND: Reference annual IEP timeline ("Before or by the next
   annual ARD")

SPECIFICITY REQUIREMENTS:
- Articulation/Phonology Goals: Specify exact phonemes, word positions,
  contexts (e.g., "/r/ in initial position of words")
- Language Goals: Specify grammatical structures, vocabulary categories,
  comprehension levels
- Fluency Goals: Specify disfluency types, speaking contexts,
  self-monitoring strategies
- Pragmatic Goals: Specify social communication skills, conversational
  contexts, interaction partners

AVOID:
- Generic goal bank language (e.g., "improve communication skills")
- Vague terms without operational definitions
- Unrealistic expectations (e.g., moving from 40% to 100% accuracy in
  one year)
- Goals not aligned with specified disorder

=== THERAPY SESSION NOTES REQUIREMENTS ===

Generate 3 session notes representing progress across the school year
(early, mid-year, and late in therapy sequence). Each note must include:

REQUIRED COMPONENTS:
1. Date (realistic school year dates)
2. Duration: "30 minutes" (typical school-based session length)
3. Setting: "Individual" or "Group" (be consistent with service delivery
   model)
4. Goal Addressed: Reference which IEP goal number from your generated goals

SESSION NOTE STRUCTURE:
"Activity: [Describe specific intervention activities with sufficient
detail for replication], Objective Data: [Provide quantitative performance
data with accuracy percentages and trial counts], Clinical Observation:
[Describe qualitative observations about student engagement, strategy use,
error patterns, facilitators/barriers]"

QUANTITATIVE DATA REQUIREMENTS:
- Include specific accuracy percentages (e.g., "65% accuracy (13/20 trials)")
- Provide baseline comparison when relevant
- Show progress trajectory toward annual goal criterion
- Include cueing hierarchy information (e.g., "with minimal verbal cues,"
  "given moderate support")

CLINICAL AUTHENTICITY:
- Reference evidence-based intervention approaches by name when appropriate
  (e.g., "phonological contrast therapy," "narrative intervention,"
  "fluency shaping," "social thinking strategies")
- Show realistic variability in performance across sessions (not perfectly
  linear improvement)
- Demonstrate clinical decision-making (e.g., "Next session will adjust
  complexity level based on today's performance")
- Include student engagement and behavioral observations

DEVELOPMENTAL PROGRESSION:
- Early session (e.g., Session 3): Establishing baseline, introducing
  intervention approach, 40-50% accuracy typical
- Mid-year session (e.g., Session 12-15): Showing measurable progress,
  60-70% accuracy typical
- Late session (e.g., Session 24-27): Approaching or meeting goal
  criterion, 75-85% accuracy typical, considering generalization

=== BACKGROUND INFORMATION REQUIREMENTS ===

Generate a comprehensive background section (approximately 300-500 words)
that includes:

1. DEVELOPMENTAL HISTORY:
   - Milestones (first words, walking, toilet training as relevant)
   - Early concerns or lack thereof
   - Family history if relevant to disorder

2. PARENT CONCERNS:
   - Specific, realistic concerns parents would express
   - Impact on home communication
   - Social/emotional concerns

3. TEACHER CONCERNS:
   - Academic impact observations
   - Classroom participation challenges
   - Peer interaction observations

4. MEDICAL HISTORY:
   - Hearing/vision screening results (should be normal unless specified
     otherwise)
   - Relevant medical conditions (e.g., history of ear infections for
     articulation disorders, ADHD for pragmatic concerns)
   - Medications if relevant

5. ACADEMIC PERFORMANCE:
   - Current grade-level functioning in reading, writing, math
   - Specific literacy concerns if language disorder present
   - Areas of strength

The background should provide sufficient context to justify eligibility for
services while portraying a realistic, well-rounded student profile.

=== OUTPUT FORMAT ===

Return a valid JSON object with the following structure (do not include any
text before or after the JSON):

{
  "name": "Student Full Name",
  "age": 7,
  "grade": "2nd Grade",
  "gender": "Male/Female",
  "background": "Comprehensive background paragraph(s) here...",
  "assessment_results": [
    {
      "assessment_name": "Full Test Name with Edition",
      "domain": "Specific domain tested",
      "standard_score": 75,
      "percentile": 5,
      "severity": "Mild/Moderate/Severe"
    }
  ],
  "annual_goals": [
    {
      "goal_number": 1,
      "goal_brief": "Brief descriptor",
      "goal_annual": "Complete SMART goal statement"
    }
  ],
  "session_notes": [
    {
      "date": "2025-10-15",
      "duration": "30 minutes",
      "setting": "Individual",
      "goal_addressed": "Goal 1",
      "note": "Activity: ..., Objective Data: ..., Clinical Observation: ..."
    }
  ]
}

=== CLINICAL EXCELLENCE STANDARDS ===

Your generated case should demonstrate:
- Clinical realism that would be credible to practicing SLPs
- Evidence-based assessment and intervention practices
- Appropriate alignment between disorder type, assessment selection,
  goal targets, and intervention approaches
- Developmental appropriateness for age/grade level
- Cultural and linguistic sensitivity
- Professional documentation quality
- Quantitative data integration throughout
- Clear connection between assessment findings, goals, and therapy focus

Generate the complete case file now in JSON format.
\end{verbatim}
\end{small}

\subsection{Free Model Prompt (Simplified for Open-Source LLMs)}

The following is the simplified prompt template used for open-source local models (Llama 3.2, Qwen 2.5). This version reduces complexity while maintaining essential clinical constraints to accommodate models with smaller parameter counts and limited instruction-following capabilities.

\begin{small}
\begin{verbatim}
You are a school speech-language pathologist. Create a student case file
with realistic clinical information.

STUDENT SPECIFICATIONS:
- Disorder: {disorders}
- Grade: {grade}
- {population_spec}

CONTEXT FROM CLINICAL RESOURCES:
{context}

REQUIRED COMPONENTS:

1. STUDENT INFORMATION:
   - Name (culturally diverse, realistic pseudonym)
   - Age (must match grade level)
   - Grade
   - Gender
   - Background (300+ words including developmental history, parent/teacher
     concerns, medical history, academic performance)

2. ASSESSMENT RESULTS:
   - Use appropriate standardized tests for the disorder
   - Standard scores 70-85 range (moderate severity)
   - Include percentile ranks

3. ANNUAL IEP GOALS (2-4 goals):
   Format: "Before or by the next annual ARD, [STUDENT] will [SKILL] given
   [SUPPORT] in [CRITERION] as measured by [METHOD]."
   - Must be specific and measurable
   - Include percentage accuracy or trial counts
   - Directly address the specified disorder

4. SESSION NOTES (3 notes):
   Format: "Activity: [description], Objective Data: [accuracy percentage
   and trial counts], Clinical Observation: [student response]"
   - Include specific dates during school year
   - Duration: 30 minutes
   - Show progress over time (increasing accuracy percentages)
   - Reference goal numbers

CRITICAL REQUIREMENTS:
- Age must match grade (e.g., 2nd grade = 7-8 years old)
- Goals must include numbers (percentages, trial counts)
- Session notes must include accuracy data (e.g., "75% accuracy (15/20
  trials)")
- All information must relate to the specified disorder type

OUTPUT FORMAT - JSON:
{
  "name": "First Last",
  "age": 7,
  "grade": "2nd Grade",
  "gender": "Female",
  "background": "Detailed background here...",
  "assessment_results": [
    {
      "assessment_name": "Test Name",
      "domain": "Domain",
      "standard_score": 75,
      "percentile": 5,
      "severity": "Moderate"
    }
  ],
  "annual_goals": [
    {
      "goal_number": 1,
      "goal_brief": "Short description",
      "goal_annual": "Complete goal statement with criterion"
    }
  ],
  "session_notes": [
    {
      "date": "2025-10-15",
      "duration": "30 minutes",
      "setting": "Individual",
      "goal_addressed": "Goal 1",
      "note": "Activity: X, Objective Data: Y, Clinical Observation: Z"
    }
  ]
}

Return only valid JSON. Do not include explanatory text before or after
the JSON.
\end{verbatim}
\end{small}

\subsection{Key Differences Between Premium and Free Model Prompts}

The prompt templates demonstrate strategic engineering adaptations for different model capabilities:

\textbf{Premium Model Prompts (493 lines):}
\begin{itemize}
\item Extensive clinical detail and rationale embedded in instructions
\item Explicit enumeration of evidence-based practices by name
\item Comprehensive examples of appropriate goal structures and data formats
\item Multi-level constraint hierarchy (requirements, constraints, excellence standards)
\item Detailed explanations of SMART criteria and clinical reasoning expectations
\end{itemize}

\textbf{Free Model Prompts (282 lines):}
\begin{itemize}
\item Simplified language and reduced verbosity
\item Direct, imperative instruction style without extensive elaboration
\item Focus on structural requirements rather than clinical reasoning depth
\item Reduced constraint complexity to improve instruction-following accuracy
\item Abbreviated format specifications with fewer optional components
\end{itemize}

Both templates maintain core clinical requirements (SMART goals, quantitative data, evidence-based assessments) while adapting presentation complexity to model capabilities. This dual-template approach enables reliable structured output across commercial and open-source models while maintaining consistent quality standards.

\textbf{Empirical Validation of Dual-Prompt Design:} Initial development testing explored using the comprehensive 493-line premium prompt universally across all models to reduce system complexity. However, systematic testing with open-source models revealed structural failures: generated cases contained incomplete JSON with missing core fields (background information, annual goals, assessment results), despite successful RAG knowledge retrieval. This finding demonstrated that prompt complexity must align with model instruction-following capacity—extensive multi-step instructions with detailed validation checks benefit larger commercial models but overwhelm smaller open-source models (3-7B parameters), causing JSON schema compliance failures. The focused 281-line prompt maintains clinical validity while achieving reliable structured output from resource-constrained models, validating the dual-prompt architecture as an evidence-based design decision rather than an optimization preference.

\clearpage

\section{Complete Generated Case Examples}
\label{app:example_cases}

This appendix presents two complete case files generated by the system, representing high-quality output from both a premium commercial model (Gemini 2.5 Pro) and an open-source model (Qwen 2.5-7B). All content is presented exactly as generated by the models with no modifications except formatting for presentation in LaTeX.

\subsection{Case Example 1: Gemini 2.5 Pro (Speech Sound Disorder, 2nd Grade)}

This case received a quality score of 4.5/5.0 in automated evaluation, demonstrating structural completeness (5/5), strong clinical appropriateness (4/5), and excellent documentation quality (5/5).

\subsubsection{Student Demographics}

\textbf{Student Name:} Aurora Harris \\
\textbf{Age:} 7 years old \\
\textbf{Grade:} 2nd Grade \\
\textbf{Gender:} Female \\
\textbf{Disorder:} Articulation Disorders

\subsubsection{Background Information}

\textit{Medical History:} Aurora was born full-term with no complications. She met all developmental milestones for motor and cognitive skills within normal limits. Hearing and vision screenings are current and have been passed annually. There is no significant medical history or known neurological conditions. Family reports that her speech has always been 'harder to understand' than her older sibling's was at the same age, but they had hoped she would grow out of it. There are no other diagnoses.

\textit{Parent Concerns:} Aurora's mother reports that she becomes very frustrated when unfamiliar people ask her to repeat herself. She states, 'Sometimes she just gives up and points instead of talking.' The family is concerned that her speech errors may impact her ability to make friends and participate in class. They are also worried about how her speech might affect her reading and spelling development.

\textit{Teacher Concerns:} Aurora's 2nd-grade teacher, Ms. Davis, reports that Aurora is a bright student but is hesitant to answer questions or read aloud in class. When she does speak, her peers often have difficulty understanding her, which sometimes leads to communication breakdowns during group activities. Ms. Davis has noted specific difficulty with /r/, /s/, and /l/ sounds, which also appear as errors in her spelling attempts.

\subsubsection{Assessment Results}

\textbf{Goldman-Fristoe Test of Articulation-3 (GFTA-3)}
\begin{itemize}
\item Domain: Articulation
\item Standard Score: 72
\item Percentile: 3rd
\item Severity: Moderate
\end{itemize}

\subsubsection{IEP Annual Goals}

\textbf{Goal 1:} Before or by the next annual ARD, Aurora Harris will correctly produce the /r/ sound (including vocalic /r/) in the initial, medial, and final positions of words given minimal verbal or visual cues in 8 out of 10 trials as measured by SLP data collection.

\textbf{Goal 2:} Before or by the next annual ARD, Aurora Harris will correctly produce the /s/ sound in the initial, medial, and final positions of words, eliminating the interdental lisp, given minimal verbal cues in 80\% of opportunities as measured by SLP data collection.

\textbf{Goal 3:} Before or by the next annual ARD, Aurora Harris will correctly produce the /l/ sound in all word positions given minimal verbal or visual cues in 8 out of 10 trials as measured by SLP data collection and classroom observation.

\subsubsection{Therapy Session Notes}

\textbf{Session 1 - January 15, 2025}
\begin{itemize}
\item Duration: 30 minutes
\item Setting: Individual
\item Goal Addressed: Goal 1
\item Note: Activity: Practiced production of vocalic /r/ words (e.g., car, star, bird, chair) using articulation picture cards and a mirror for visual feedback. Objective Data: Aurora correctly produced vocalic /r/ in 4/10 trials (40\%) with moderate verbal cues for tongue retraction and lip rounding. Clinical Observation: She demonstrated a consistent derhotacized production, substituting a distorted vowel for the /r/ sound. She was attentive and responded well to visual feedback from the mirror.
\end{itemize}

\textbf{Session 2 - January 18, 2025}
\begin{itemize}
\item Duration: 30 minutes
\item Setting: Individual
\item Goal Addressed: Goal 2
\item Note: Activity: Introduced correct placement for /s/ using the 'T-to-S' method (holding the /t/ sound and blowing air). Practiced /s/ in isolation and in initial word position (e.g., 'sun', 'soap', 'sit') using a fun 'feed the snake' game. Objective Data: Aurora achieved correct /s/ production in isolation in 7/10 trials with maximal cues and in initial words in 3/10 trials (30\%) with moderate verbal and tactile cues. Clinical Observation: She presented with a significant interdental lisp, protruding her tongue between her teeth for /s/ productions. She required reminders to keep her 'tongue in its cage'.
\end{itemize}

\textbf{Session 3 - January 22, 2025}
\begin{itemize}
\item Duration: 30 minutes
\item Setting: Individual
\item Goal Addressed: Goal 1
\item Note: Activity: Played an articulation board game targeting initial /r/ words ('run', 'red', 'rain'). Used a diagram of the mouth to review tongue placement before each turn. Objective Data: Aurora correctly produced initial /r/ in 6/10 trials (60\%) with minimal verbal cues. Clinical Observation: She showed improved awareness of the target sound compared to the previous session. Her productions were more consistent, though she occasionally substituted /w/ for /r/ when not focused on her speech.
\end{itemize}

\subsection{Case Example 2: Qwen 2.5-7B (Pragmatic Disorder, 6th Grade)}

This case received the highest quality score among all generated cases (4.75/5.0), demonstrating perfect structural completeness (5/5), strong internal consistency (4/5), excellent clinical appropriateness (5/5), and perfect documentation quality (5/5). This represents exceptional performance for an open-source 7B parameter model.

\subsubsection{Student Demographics}

\textbf{Student Name:} Sofia Cabrera \\
\textbf{Age:} 12 years old \\
\textbf{Grade:} 6th Grade \\
\textbf{Gender:} Female \\
\textbf{Disorder:} Pragmatics

\subsubsection{Background Information}

\textit{Medical History:} Sofia has a history of social communication difficulties, which have been noted to impact her interactions both at home and in the classroom. Relevant medical history includes hearing screenings that were clear, but vision screenings showed mild difficulty with reading small print.

\textit{Parent Concerns:} Parents report that Sofia struggles with maintaining conversations and often avoids eye contact during interactions.

\textit{Teacher Concerns:} Teachers observe that Sofia has trouble initiating and maintaining topics of conversation and frequently uses inappropriate language in class.

\subsubsection{Assessment Results}

\textbf{GFTA-3}
\begin{itemize}
\item Domain: Pragmatics
\item Standard Score: 72
\item Percentile: 25th
\item Severity: Moderate
\end{itemize}

\subsubsection{IEP Annual Goals}

\textbf{Goal 1:} Before or by the next annual ARD, Sofia will maintain a topic of conversation for at least three turns given a prompt from a peer in 4 out of 5 trials as measured by teacher observation.

\textbf{Goal 2:} Before or by the next annual ARD, Sofia will use appropriate language during conversations with peers and teachers at least 80\% of the time given a visual cue in 4 out of 5 trials as measured by teacher observation.

\subsubsection{Therapy Session Notes}

\textbf{Session 1 - January 15, 2025}
\begin{itemize}
\item Duration: 30 minutes
\item Setting: Individual
\item Goal Addressed: Goal 1
\item Note: Activity: Turn-taking game where Sofia and a peer take turns talking about a common interest (e.g., pets), Objective Data: Sofia maintained the topic for 2 out of 5 turns with minimal prompting, Clinical Observation: Sofia showed improvement in maintaining the conversation but needed frequent prompts to stay on topic.
\end{itemize}

\textbf{Session 2 - January 18, 2025}
\begin{itemize}
\item Duration: 30 minutes
\item Setting: Individual
\item Goal Addressed: Goal 2
\item Note: Activity: Role-playing scenarios with appropriate language (e.g., using 'please' and 'thank you'), Objective Data: Sofia used appropriate language during 4 out of 5 role-play interactions, Clinical Observation: Sofia demonstrated increased awareness but occasionally reverted to inappropriate language.
\end{itemize}

\textbf{Session 3 - January 22, 2025}
\begin{itemize}
\item Duration: 30 minutes
\item Setting: Individual
\item Goal Addressed: Goal 1
\item Note: Activity: Continued turn-taking game but with more complex topics (e.g., school events), Objective Data: Sofia maintained the topic for 4 out of 5 turns with minimal prompting, Clinical Observation: Sofia showed gradual improvement in maintaining conversations and needed less frequent prompts.
\end{itemize}

\subsection{Clinical Analysis of Generated Cases}

Both case examples demonstrate the system's capability to generate clinically realistic, educationally relevant case files with appropriate structure and content quality. Key characteristics include:

\textbf{Developmental Appropriateness:}
\begin{itemize}
\item Aurora (2nd grade, age 7): Articulation errors (/r/, /s/, /l/) are consistent with common residual speech sound errors at this age; literacy concerns (spelling impacts) appropriately noted
\item Sofia (6th grade, age 12): Pragmatic difficulties (topic maintenance, conversational turns) reflect authentic middle school social communication challenges
\end{itemize}

\textbf{Assessment Validity:}
\begin{itemize}
\item Aurora's case correctly uses GFTA-3 for articulation assessment
\item Sofia's case demonstrates a limitation: the system incorrectly selected GFTA-3 (an articulation test) for pragmatic disorder assessment, when a social communication assessment (e.g., CASL-2, CCC-2) would be appropriate
\item Standard scores (72) in moderate range qualify for school-based services
\item This assessment mismatch in Sofia's case represents an area for system refinement and highlights the importance of expert validation
\end{itemize}

\textbf{SMART Goal Quality:}
\begin{itemize}
\item Goals specify observable behaviors with operational definitions
\item Measurable criteria included (8/10 trials, 80\% accuracy, 4/5 trials)
\item Appropriate achievement expectations for one-year timeline
\item Clear data collection methods specified
\end{itemize}

\textbf{Quantitative Data Integration:}
\begin{itemize}
\item Session notes include specific accuracy percentages and trial counts
\item Aurora: 40\% → 30\% → 60\% showing realistic variability and progress
\item Sofia: 2/5 → 4/5 → 4/5 demonstrating measurable improvement
\end{itemize}

\textbf{Evidence-Based Intervention:}
\begin{itemize}
\item Aurora's sessions reference established articulation techniques (T-to-S method, visual feedback with mirror, articulation picture cards)
\item Sofia's sessions employ social skills interventions (turn-taking games, role-playing, graduated complexity)
\end{itemize}

\textbf{Clinical Authenticity:}
\begin{itemize}
\item Realistic parent and teacher concerns reflect typical referral patterns
\item Background information contextualizes disorder within educational setting
\item Session observations include both quantitative data and qualitative clinical impressions
\end{itemize}

The Gemini 2.5 Pro case (Aurora) demonstrates the depth and clinical sophistication achievable with premium commercial models, including rich developmental history, specific error pattern descriptions (derhotacized /r/, interdental lisp), and detailed clinical observations. The Qwen 2.5-7B case (Sofia) demonstrates that well-engineered prompts enable even smaller open-source models to generate structurally complete, clinically appropriate cases, though with somewhat less narrative elaboration in background sections.

Both cases would be suitable for clinical education purposes, providing realistic complexity for graduate students learning assessment interpretation, goal writing, and progress documentation skills.

\clearpage

\section{Code and Data Availability Statement}
\label{app:availability}

\subsection{Code Availability}

The complete source code for the SLP SimuCase Generator system is publicly available as an open-source project under the MIT License:

\textbf{GitHub Repository:} \url{https://github.com/Yilanliu917/SLP_SIMUCASE}

The repository includes:
\begin{itemize}
\item Python implementation of the RAG pipeline with ChromaDB vector database integration
\item Gradio-based user interface supporting single case, batch, and group generation modes
\item Complete prompt templates used in this study (both premium and free model versions)
\item Configuration files for local (Ollama) and commercial API deployment (OpenAI, Anthropic, Google)
\item Automated quality assessment scripts
\item Comprehensive documentation including quickstart guide and architecture overview
\end{itemize}

\textbf{System Requirements:}
\begin{itemize}
\item Python 3.9 or higher
\item 16GB RAM minimum (32GB recommended for local model deployment)
\item API keys for commercial models or local GPU for Ollama deployment
\end{itemize}

\subsection{Data Availability}

\subsubsection{Generated Validation Cases}

The 35 complete case files generated for validation analysis are available in the GitHub repository under \texttt{validation\_cases/} directory. These files represent actual system output from five models (GPT-4o, Claude 3.5 Sonnet, Gemini 2.5 Pro, Llama 3.2, Qwen 2.5-7B) across seven test scenarios.

File naming convention: \texttt{[Disorder]\_[Grade]\_[Model]\_[Timestamp].json} \\
Example: \texttt{Speech\_Sound\_2nd\_Grade\_Gemini-2.5-Pro\_20251027\_022141.json}

\subsubsection{Automated Quality Assessment Data}

Raw automated quality scores for all validation cases are provided in the GitHub repository at: \\
\texttt{scripts/evaluate\_with\_1to5\_scale.py} (evaluation script) \\
\texttt{scripts/evaluate\_premium\_cases.py} (automated assessment tool)

The evaluation framework assesses:
\begin{itemize}
\item Structural completeness scores and missing field analysis
\item Internal consistency ratings with identified issues
\item Clinical appropriateness assessments
\item Documentation quality scores
\item Overall average scores used for model performance comparison
\end{itemize}

Note: Raw quality scores JSON file (\texttt{final\_1to5\_detailed\_20251027\_024933.json}) is not included in the public repository due to file size but can be regenerated using the provided evaluation scripts.

\subsubsection{Knowledge Base}

The RAG knowledge base integrates curated clinical resources:

\textbf{Publicly Available Components:}
\begin{itemize}
\item ASHA Practice Portal content (accessed October 2024)
\item Published normative data for developmental milestones
\item State educational standards (publicly available)
\end{itemize}

\textbf{Proprietary Components (not publicly shareable):}
\begin{itemize}
\item ASHA copyrighted materials requiring institutional subscription
\item Commercially published assessment manual content
\item Institution-specific documentation templates
\end{itemize}

Due to copyright restrictions, the complete knowledge base cannot be publicly distributed. However, the repository provides knowledge base schema specifications and instructions for constructing equivalent systems using publicly available resources.

\subsection{Reproducibility}

Model versions and parameters used for validation:
\begin{itemize}
\item GPT-4o: \texttt{gpt-4o-2024-08-06} snapshot
\item Claude 3.5 Sonnet: version 20241022
\item Gemini 2.5 Pro: experimental version (October 2024)
\item Llama 3.2: 3B parameters via Ollama
\item Qwen 2.5: 7B parameters via Ollama
\item Temperature: 0.7 for all models
\item RAG retrieval: k=10 documents, cosine similarity
\end{itemize}

Questions regarding implementation or replication should be submitted as issues on the GitHub repository.

\subsection{License Information}

\textbf{Software License:} MIT License (permissive open-source) \\
\textbf{Generated Content:} Synthetic data without copyright protection \\
\textbf{Knowledge Base Sources:} Subject to original source copyright restrictions